\documentclass{article} 
\usepackage{iclr2017_workshop,times}
\usepackage{hyperref}
\usepackage{url}
\usepackage{graphicx}
\usepackage{epstopdf}
\usepackage{fouridx} 
\usepackage{amsmath}

\title{Trace Norm Regularised Deep Multi-Task Learning}

\author{Yongxin Yang, Timothy M. Hospedales\\
Queen Mary, University of London\\
\texttt{\{yongxin.yang, t.hospedales\}@qmul.ac.uk} \\
}

\begin{document}

\maketitle

\begin{abstract}
We propose a framework for training multiple neural networks simultaneously. The parameters from all models are regularised by the tensor trace norm, so that each neural network is encouraged to reuse others' parameters if possible -- this is the main motivation behind multi-task learning. In contrast to many deep multi-task learning models, we do not predefine a parameter sharing strategy by specifying which layers have tied parameters. Instead, our framework considers sharing for all shareable layers, and the sharing strategy is learned in a data-driven way. 
\end{abstract}

\section{Introduction and Related Work}

Multi-task learning (MTL) \citep{Caruana1997} aims to learn multiple tasks jointly, so that knowledge obtained from one task can be reused by others. We first briefly review some studies in this area.

\vspace{0.1cm}\noindent\textbf{Matrix-based Multi-Task Learning}\quad
Matrix-based MTL is usually built on linear models, i.e., each task is parameterised by a $D$-dimensional weight vector $w$, and the model prediction is $\hat{y} = x \cdot w = x^T w$, where $x$ is a $D$-dimensional feature vector representing an instance. The objective function for matrix-based MTL can be written as $\sum_{i=1}^{T}\sum_{j=1}^{N^{(i)}} \ell(y^{(i)}_j, x^{(i)}_j \cdot w^{(i)}) + \lambda \Omega(W)$. Here $\ell(y, \hat{y})$ is a loss function of the true label $y$ and  predicted label $\hat{y}$. $T$ is the number of tasks, and for the $i$-th task there are $N^{(i)}$ training instances. Assuming the dimensionality of every task's feature is the same, the models -- $w^{(i)}$s -- are of the same size. Then the collection of $w^{(i)}$s forms a $D\times T$ matrix $W$ of which the $i$-th column is the linear model for the $i$-t task. To achieve MTL we exploit a regulariser $\Omega(W)$ that couples the learning problems, typically by encouraging $W$ to be a low-rank matrix. Some choices include the $\ell_{2,1}$ norm \citep{Argyriou2008}, and trace norm \citep{Ji2009tracenormmtl}. An alternative approach \citep{daume2012gomtl} is to explicitly formulate $W$ as a low-rank matrix, i.e., $W=LS$ where $L$ is a $D\times K$ matrix and $S$ is a $K\times T$ matrix with $K<\min(D,T)$ as a hyper-parameter (matrix rank).  

\vspace{0.1cm}\noindent\textbf{Tensor-based Multi-Task Learning}\quad
In the classic MTL setting, each task is indexed by a single factor. But in many real-world problems, tasks are indexed by multiple factors. For example, to build a restaurant recommendation system, we want a regression model that predicts the scores for different aspects (food quality, environment) by different customers. Then the task is indexed by aspects $\times$ customers. The collection of linear models for all tasks is then a 3-way tensor $\mathcal{W}$ of size $D\times T_1 \times T_2$, where $T_1$ and $T_2$ is the number of aspects and customers respectively. Consequently $\Omega(\mathcal{W})$ has to be a  tensor regulariser \citep{Tomioka10onthe}. For example, sum of the trace norms on all matriciations\footnote{Matriciation is also known as tensor unfolding or flattening.} \citep{icml2013_romera-paredes13}, and scaled latent trace norm \citep{wimalawarnemultitask}. An alternative solution is to concatenate the one-hot encodings of the two task factors and feed it as input into a two-branch neural network model \citep{yang15}.

\vspace{0.1cm}\noindent\textbf{Multi-Task Learning for Neural Networks}\quad
With the success of deep learning, many studies have investigated deep multi-task learning. \cite{ZhangLLT14} use a convolutional neural network to find facial landmarks as well as recognise face attributes (e.g., emotions). \cite{Liu15nlp} propose a  neural network for query classification and information retrieval (ranking for web search). A key commonality of these studies is that they  use a predefined  sharing strategy. A typical design is to use the same parameters for the bottom layers of the deep neural network and task-specific parameters for the top layers. This kind of architecture  can be traced back to 2000s \citep{bak02b}. However, modern neural network architectures  contain a large number of layers, which makes the decision of `\textit{at which layer to split the neural network for different tasks?}' extremely hard.

\section{Methodology}

Instead of predefining a parameter sharing strategy, we propose the following framework: For $T$ tasks, each is modelled by a neural network of the same architecture. 
We collect the parameters in a layer-wise fashion, and put a tensor norm on every collection.
We illustrate the idea by a simple example: assume that we have $T=2$ tasks, and each is modelled by a $4$-layer convolution neural network (CNN). The CNN architecture is: (1) convolutional layer (`conv1') of size $5\times5\times3\times32$, (2) `conv2' of size $3\times3\times32\times64$, (3) fully-connected layer (`fc1') of size $256\times256$, (4) fully-connected layer `fc2'$^{(1)}$ of size $256\times10$ for the first task and fully-connected layer (`fc2'$^{(2)}$) of size $256\times20$ for the second task. Since the two tasks have different numbers of outputs, the potentially shareable layers are `conv1', `conv2', and `fc1', excluding the final layer of different dimensionality.

For single task learning, the parameters are `conv1'$^{(1)}$, `conv2'$^{(1)}$, `fc1'$^{(1)}$, and `fc2'$^{(1)}$ for the first task; `conv1'$^{(2)}$, `conv2'$^{(2)}$, `fc1'$^{(2)}$, and `fc2'$^{(2)}$ for the second task. We can see that there is not any parameter sharing between these two tasks.
In one possible predefined deep MTL architecture, the parameters could be `conv1', `conv2', `fc1'$^{(1)}$, and `fc2'$^{(1)}$ for the first task; `conv1', `conv2', `fc1'$^{(2)}$, and `fc2'$^{(2)}$ for the second task, i.e., the first and second layer are fully shared in this case.
For our proposed method, the parameter setting is the same as single task learning mode, but we put three tensor norms on the stacked \{`conv1'$^{(1)}$, `conv1'$^{(2)}$\} (a tensor of size $5\times5\times3\times32\times2$), the stacked \{`conv2'$^{(1)}$, `conv2'$^{(2)}$\} (a tensor of size $3\times3\times32\times64\times2$), and the stacked \{`fc1'$^{(1)}$, `fc1'$^{(2)}$\} (a tensor of size $256\times256\times2$) respectively.

\vspace{0.1cm}\noindent\textbf{Tensor Norm}\quad
We choose to use the trace norm,  the sum of a matrix's singular values $||X||_* = \sum_{i=1} \sigma_i$. It has a nice property that it is the tightest convex relation of matrix rank \citep{Recht2010}. When directly restricting the rank of a matrix is challenging, trace norm serves as a good proxy. The extension of trace norm from matrix to tensor is not unique, just like tensor rank has multiple definitions. How to define tensor rank depends on how we assume the tensor is factorised, e.g., Tucker  \citep{Tuck1966c} and Tensor-Train  \cite{Oseledets2011} decompositions. We propose three tensor trace norm designs here, which  correspond to three variants of the proposed method.

For an $N$-way tensor $\mathcal{W}$ of size $D_1\times D_2 \times \dots \times D_N$. We define
\begin{equation}
\textbf{(Tensor Trace Norm) Last Axis Flattening} \quad ||\mathcal{W}||_* = \gamma ||\mathcal{W}_{(N)}||_*
\label{eq:mtxmtl}
\end{equation}
$\mathcal{W}_{(i)} := \operatorname{reshape}(\operatorname{permute}(\mathcal{W}, [i, 1, \dots, i-1, i+1 \dots, N]), [D_i, \prod_{j\neg i} D_j])$ is the mode-$i$ tensor flattening. This is the simplest definition. Given that in our framework, the last axis of tensor indexes the tasks, i.e., $D_N = T$, it is the most straightforward way to adapt the technique of matrix-based MTL -- reshape the $D_1\times D_2 \times \dots \times T$ tensor to $D_1 D_2 \dots \times T$ matrix. 

To advance, we define two kinds of tensor trace norm that are closely connected with Tucker-rank (obtained by Tucker decomposition) and TT-rank (obtained by Tensor Train decomposition).
\begin{eqnarray}
\textbf{(Tensor Trace Norm) Tucker} \quad ||\mathcal{W}||_* &=& \sum_{i=1}^{N} \gamma_i ||\mathcal{W}_{(i)}||_*
\label{eq:tuckermtl}\\
%
\textbf{(Tensor Trace Norm) TT} \quad ||\mathcal{W}||_* &=& \sum_{i=1}^{N-1} \gamma_i ||\mathcal{W}_{[i]}||_*
\label{eq:ttmtl}
\end{eqnarray}

Here $\mathcal{W}_{[i]}$ is yet another way to unfold the tensor, which is obtained by $\mathcal{W}_{[i]} = \operatorname{reshape}(\mathcal{W}, [D_1 D_2 \dots D_i, D_{i+1} D_{i+2} \dots D_N])$. It is interesting to note that unlike LAF, Tucker and TT also encourage within-task parameter sharing, e.g, sharing across filters in a neural network context.

\vspace{0.1cm}\noindent\textbf{Optimisation}\quad
Using gradient-based methods for optimisation involving trace norm is \emph{not} a common choice, as there are better solutions based on semi-definite programming or proximal gradients since the trace norm is essentially non-differentiable. However, deep neural networks are usually trained by gradient descent, and we prefer to keep the standard training process. Therefore we use (sub-)gradient descent. The sub-gradient for trace norm can be derived as $\frac{\partial ||X||_*}{\partial X} = X(X^T X)^{-\frac{1}{2}}$. A more numerical stable method instead of computing the inverse matrix square root is $X(X^T X)^{-\frac{1}{2}} = U V^T$ where $U$ and $V$ are obtained from SVD: $X=U\Sigma V^T$ \citep{Watson1992Characterization}.

\section{Experiment}

Our method is implemented in TensorFlow \citep{tensorflow2015}, and released on Github\footnote{\url{https://github.com/wOOL/TNRDMTL}}. 
We experiment on the Omniglot dataset \citep{Lake1332}. Omniglot contains handwritten letters in 50 different alphabets (e.g., Cyrillic, Korean, Tengwar),  each with its own number of unique characters ($14\sim 55$). In total, there are 1623 unique characters, each with 20 instances. Each task is a multi-class character recognition problem for the corresponding alphabet. 
The images are monochrome of size $105\times 105$. We design a CNN  with $3$ convolutional  and $2$ FC layers. The first conv layer has $8$ filters of size $5\times 5$; the second conv layer has $12$ filters of size $3\times 3$, and the third convolutional layer has $16$ filters of size $3\times 3$. Each convolutional layer is followed by a $2\times 2$ max-pooling. The first FC layer has $64$ neurons, and the second FC layer has size corresponding to the number of unique classes in the alphabet. The activation function is $tanh$.
We compare the three variants of the proposed framework -- LAF (Eq.~\ref{eq:mtxmtl}), Tucker (Eq.~\ref{eq:tuckermtl}), and TT (Eq.~\ref{eq:ttmtl}) with single task learning (STL). For every layer, there are one (LAF) or more (Tucker and TT) $\gamma$ that control the trade-off between the classification loss (cross-entropy) and the trace norm terms, for which we set all $\gamma=0.01$.

\begin{figure}[t]
	\centering
	\includegraphics[width=0.95\linewidth]{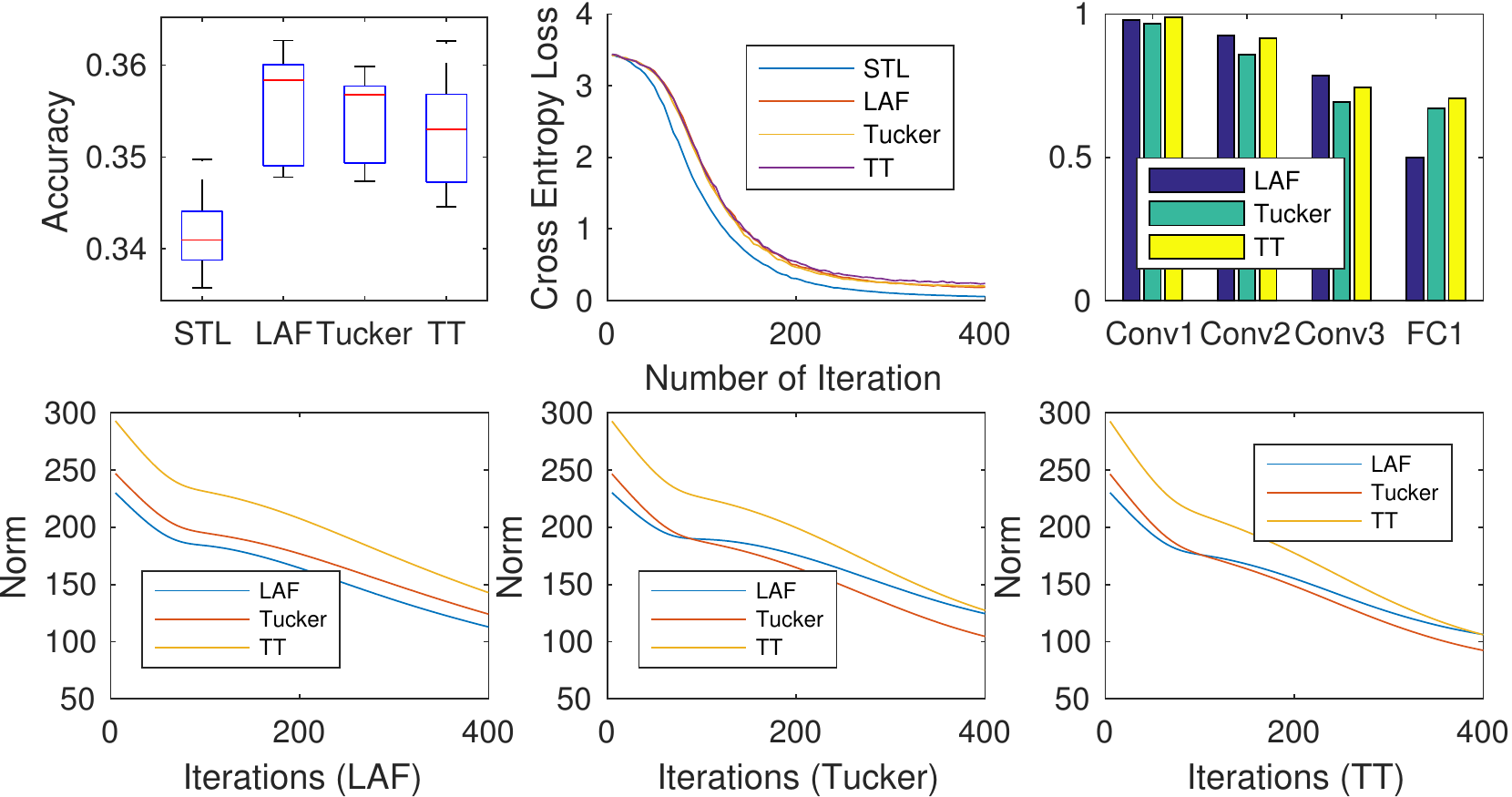}
	\caption{Top-left: Testing accuracy. Top-mid: Training loss. Top-right: sharing strength by layer. Bottom: Norms when optimising LAF (left), Tucker (middle), TT (right).}
	\label{fig:loss_plot_ce_and_norm}
	\vspace{-0.4cm}
\end{figure}

The experiments are repeated $10$ times, and every time $10\%$ training data and $90\%$ testing data are randomly selected. We plot the change of cross-entropy loss in training set and the values of norm terms with the neural networks' parameters updating. As we can see in Fig~\ref{fig:loss_plot_ce_and_norm}, STL has the lowest training loss, but worst testing performance, suggesting over-fitting. Our methods  alleviate the problem with multi-task regularisation. We roughly estimate the strength of parameter sharing by calculating $1-\frac{\text{Norm of Optimised Param}}{\text{Norm of Initialised Param}}$, we can see the pattern that with bottom layers share more compared to the top ones. This reflects the common design intuition that the bottom layers are more data/task independent. Finally, it appears that the choice on LAF, Tucker, or TT may not be very sensitive as we observe that when optimising one, the loss of the other norms still reduces. 

This technique  provides a data-driven solution to the branching architecture design problem in deep multi-task learning. It is a flexible norm regulariser-based alternative to explicit factorisation-based approaches to the same problem \citep{yang2017dmrl}.


\begin{thebibliography}{18}
	\providecommand{\natexlab}[1]{#1}
	\providecommand{\url}[1]{\texttt{#1}}
	\expandafter\ifx\csname urlstyle\endcsname\relax
	\providecommand{\doi}[1]{doi: #1}\else
	\providecommand{\doi}{doi: \begingroup \urlstyle{rm}\Url}\fi
	
	\bibitem[Abadi et~al.(2015)Abadi, Agarwal, Barham, Brevdo, Chen, Citro,
	Corrado, Davis, Dean, Devin, Ghemawat, Goodfellow, Harp, Irving, Isard, Jia,
	Jozefowicz, Kaiser, Kudlur, Levenberg, Man\'{e}, Monga, Moore, Murray, Olah,
	Schuster, Shlens, Steiner, Sutskever, Talwar, Tucker, Vanhoucke, Vasudevan,
	Vi\'{e}gas, Vinyals, Warden, Wattenberg, Wicke, Yu, and
	Zheng]{tensorflow2015}
	Mart\'{\i}n Abadi, Ashish Agarwal, Paul Barham, Eugene Brevdo, Zhifeng Chen,
	Craig Citro, Greg~S. Corrado, Andy Davis, Jeffrey Dean, Matthieu Devin,
	Sanjay Ghemawat, Ian Goodfellow, Andrew Harp, Geoffrey Irving, Michael Isard,
	Yangqing Jia, Rafal Jozefowicz, Lukasz Kaiser, Manjunath Kudlur, Josh
	Levenberg, Dan Man\'{e}, Rajat Monga, Sherry Moore, Derek Murray, Chris Olah,
	Mike Schuster, Jonathon Shlens, Benoit Steiner, Ilya Sutskever, Kunal Talwar,
	Paul Tucker, Vincent Vanhoucke, Vijay Vasudevan, Fernanda Vi\'{e}gas, Oriol
	Vinyals, Pete Warden, Martin Wattenberg, Martin Wicke, Yuan Yu, and Xiaoqiang
	Zheng.
	\newblock {TensorFlow}: Large-scale machine learning on heterogeneous systems,
	2015.
	\newblock URL \url{http://tensorflow.org/}.
	\newblock Software available from tensorflow.org.
	
	\bibitem[Argyriou et~al.(2008)Argyriou, Evgeniou, and Pontil]{Argyriou2008}
	Andreas Argyriou, Theodoros Evgeniou, and Massimiliano Pontil.
	\newblock Convex multi-task feature learning.
	\newblock \emph{Machine Learning}, 2008.
	
	\bibitem[Bakker \& Heskes(2003)Bakker and Heskes]{bak02b}
	Bart Bakker and Tom Heskes.
	\newblock Task clustering and gating for {B}ayesian multitask learning.
	\newblock \emph{Journal of Machine Learning Research (JMLR)}, 2003.
	
	\bibitem[Caruana(1997)]{Caruana1997}
	Rich Caruana.
	\newblock Multitask learning.
	\newblock \emph{Machine Learning}, 1997.
	
	\bibitem[Ji \& Ye(2009)Ji and Ye]{Ji2009tracenormmtl}
	Shuiwang Ji and Jieping Ye.
	\newblock An accelerated gradient method for trace norm minimization.
	\newblock In \emph{International Conference on Machine Learning (ICML)}, 2009.
	
	\bibitem[Kumar \& {Daum\'e III}(2012)Kumar and {Daum\'e III}]{daume2012gomtl}
	Abhishek Kumar and Hal {Daum\'e III}.
	\newblock Learning task grouping and overlap in multi-task learning.
	\newblock In \emph{International Conference on Machine Learning (ICML)}, 2012.
	
	\bibitem[Lake et~al.(2015)Lake, Salakhutdinov, and Tenenbaum]{Lake1332}
	Brenden~M. Lake, Ruslan Salakhutdinov, and Joshua~B. Tenenbaum.
	\newblock Human-level concept learning through probabilistic program induction.
	\newblock \emph{Science}, 2015.
	
	\bibitem[Liu et~al.(2015)Liu, Gao, He, Deng, Duh, and Wang]{Liu15nlp}
	Xiaodong Liu, Jianfeng Gao, Xiaodong He, Li~Deng, Kevin Duh, and Ye-Yi Wang.
	\newblock Representation learning using multi-task deep neural networks for
	semantic classification and information retrieval.
	\newblock \emph{NAACL}, 2015.
	
	\bibitem[Oseledets(2011)]{Oseledets2011}
	I.~V. Oseledets.
	\newblock Tensor-train decomposition.
	\newblock \emph{SIAM Journal on Scientific Computing}, 2011.
	
	\bibitem[Recht et~al.(2010)Recht, Fazel, and Parrilo]{Recht2010}
	Benjamin Recht, Maryam Fazel, and Pablo~A. Parrilo.
	\newblock Guaranteed minimum-rank solutions of linear matrix equations via
	nuclear norm minimization.
	\newblock \emph{SIAM Rev.}, 2010.
	
	\bibitem[Romera-paredes et~al.(2013)Romera-paredes, Aung, Bianchi-berthouze,
	and Pontil]{icml2013_romera-paredes13}
	Bernardino Romera-paredes, Hane Aung, Nadia Bianchi-berthouze, and Massimiliano
	Pontil.
	\newblock Multilinear multitask learning.
	\newblock In \emph{International Conference on Machine Learning (ICML)}, 2013.
	
	\bibitem[Tomioka et~al.(2010)Tomioka, Hayashi, and Kashima]{Tomioka10onthe}
	Ryota Tomioka, Kohei Hayashi, and Hisashi Kashima.
	\newblock On the extension of trace norm to tensors.
	\newblock In \emph{NIPS Workshop on Tensors, Kernels, and Machine Learning},
	2010.
	
	\bibitem[Tucker(1966)]{Tuck1966c}
	L.~R. Tucker.
	\newblock Some mathematical notes on three-mode factor analysis.
	\newblock \emph{Psychometrika}, 1966.
	
	\bibitem[Watson(1992)]{Watson1992Characterization}
	G.A. Watson.
	\newblock Characterization of the subdifferential of some matrix norms.
	\newblock \emph{Linear Algebra and its Applications}, 170:\penalty0 33 -- 45,
	1992.
	\newblock ISSN 0024-3795.
	\newblock \doi{http://dx.doi.org/10.1016/0024-3795(92)90407-2}.
	\newblock URL
	\url{http://www.sciencedirect.com/science/article/pii/0024379592904072}.
	
	\bibitem[Wimalawarne et~al.(2014)Wimalawarne, Sugiyama, and
	Tomioka]{wimalawarnemultitask}
	Kishan Wimalawarne, Masashi Sugiyama, and Ryota Tomioka.
	\newblock Multitask learning meets tensor factorization: task imputation via
	convex optimization.
	\newblock In \emph{Neural Information Processing Systems (NIPS)}, 2014.
	
	\bibitem[Yang \& Hospedales(2015)Yang and Hospedales]{yang15}
	Yongxin Yang and Timothy~M. Hospedales.
	\newblock A unified perspective on multi-domain and multi-task learning.
	\newblock In \emph{International Conference on Learning Representations
		(ICLR)}, 2015.
	
	\bibitem[Yang \& Hospedales(2017)Yang and Hospedales]{yang2017dmrl}
	Yongxin Yang and Timothy~M. Hospedales.
	\newblock Deep multi-task representation learning: {A} tensor factorisation
	approach.
	\newblock In \emph{International Conference on Learning Representations
		(ICLR)}, 2017.
	
	\bibitem[Zhang et~al.(2014)Zhang, Luo, Loy, and Tang]{ZhangLLT14}
	Zhanpeng Zhang, Ping Luo, Chen~Change Loy, and Xiaoou Tang.
	\newblock Facial landmark detection by deep multi-task learning.
	\newblock In \emph{European Conference on Computer Vision (ECCV)}, 2014.
	
\end{thebibliography}
\end{document}